\documentclass[sn-mathphys-num]{sn-jnl}

\usepackage{graphicx}%
\usepackage{multirow}%
\usepackage{amsmath,amssymb,amsfonts}%
\usepackage{amsthm}%
\usepackage{mathrsfs}%
\usepackage[title]{appendix}%
\usepackage{xcolor}%
\usepackage{textcomp}%
\usepackage{manyfoot}%
\usepackage{booktabs}%
\usepackage{algorithm}%
\usepackage{algorithmicx}%
\usepackage{algpseudocode}%
\usepackage{listings}%

\theoremstyle{thmstyleone}%
\theoremstyle{thmstyletwo}%

\theoremstyle{thmstylethree}%

\begin{document}

\title[Surgical-DeSAM: Decoupling SAM for Instrument Segmentation in Robotic Surgery]{Surgical-DeSAM: Decoupling SAM for Instrument Segmentation in Robotic Surgery}


\author[1,2]{\fnm{Yuyang} \sur{Sheng}}\email{yuyang.sheng.22@alumni.ucl.ac.uk}

\author[1,2]{\fnm{Sophia} \sur{Bano}}\email{sophia.bano@ucl.ac.uk}

\author[1]{\fnm{Matthew J.} \sur{Clarkson}}\email{m.clarkson@ucl.ac.uk}

\author[1]{\fnm{Mobarakol} \sur{Islam}}\email{mobarakol.islam@ucl.ac.uk} 

\affil[1]{\orgdiv{Wellcome/EPSRC Centre for Interventional and Surgical Sciences (WEISS) and Department of Medical Physics and Biomedical Engineering}, \orgname{University College London}, \orgaddress{\city{London},  \country{UK}}}

\affil[2]{\orgdiv{Dept. of Computer Science}, \orgname{University College London}, \orgaddress{\city{London},  \country{UK}}}

\abstract{\textbf{Purpose:} The recent Segment Anything Model (SAM) has demonstrated impressive performance with point, text or bounding box prompts, in various applications. However, in safety-critical surgical tasks, prompting is not possible due to (i) the lack of per-frame prompts for supervised learning, (ii) it is unrealistic to prompt frame-by-frame in a real-time tracking application, and (iii)  it is expensive to annotate prompts for offline applications.

\textbf{Methods:} We develop Surgical-DeSAM to generate automatic bounding box prompts for decoupling SAM to obtain instrument segmentation in real-time robotic surgery. We utilise a commonly used detection architecture, DETR, and fine-tuned it to obtain bounding box prompt for the instruments. We then empolyed decoupling SAM (DeSAM) by replacing the image encoder with DETR encoder and fine-tune prompt encoder and mask decoder to obtain instance segmentation for the surgical instruments. To improve detection performance, we adopted the Swin-transformer to better feature representation.

\textbf{Results:} The proposed method has been validated on two publicly available datasets from the MICCAI surgical instruments segmentation challenge EndoVis 2017 and 2018. The performance of our method is also compared with SOTA instrument segmentation methods and demonstrated significant improvements with dice metrics of 89.62 and 90.70 for the EndoVis 2017 and 2018.

\textbf{Conclusion:} Our extensive experiments and validations demonstrate that Surgical-DeSAM enables real-time instrument segmentation without any additional prompting and outperforms other SOTA segmentation methods.}


\maketitle

\section{Introduction}\label{sec:intro}
\label{sec:intro}
Robot-assist surgery is gaining increasing attention in the research field of intelligent robots. Some existing works apply deep learning techniques to realize instance segmentation for surgical instruments. While these models have significantly advanced instance segmentation performance on surgical datasets, they have yet to fully harness the capabilities of either the most recent segmentation models or the advanced object detection model, which presents an opportunity for further refinement and enhancement. The well-known segmentation foundation model, SAM (Segment Anything Model) \cite{kirillov2023segment}, and adaptations of SAM in medical image segmentation and surgical instrument segmentation \cite{ma2023segment} have shown great promise in semantic segmentation. However, they cannot produce object label segmentation, and they require interactive prompting during the deployment period, which is not realistic.

In this work, we (i) propose Surgical-DeSAM to generate automatic bounding box prompting for a decoupling SAM; (ii) design Swin-DETR by replacing ResNet with Swin-transformer as image feature extractor of the DETR~\cite{carion2020end}; (iii) decouple SAM (DeSAM) by replacing SAM's image encoder with DETR's encoder; (iv) validate on two publicly available surgical instrument segmentation datasets of EndoVis17 and EndoVis18; and (v) demonstrate the robustness compared to the SOTA models.

\begin{figure}[!h]
\centering
\includegraphics[scale=.4]{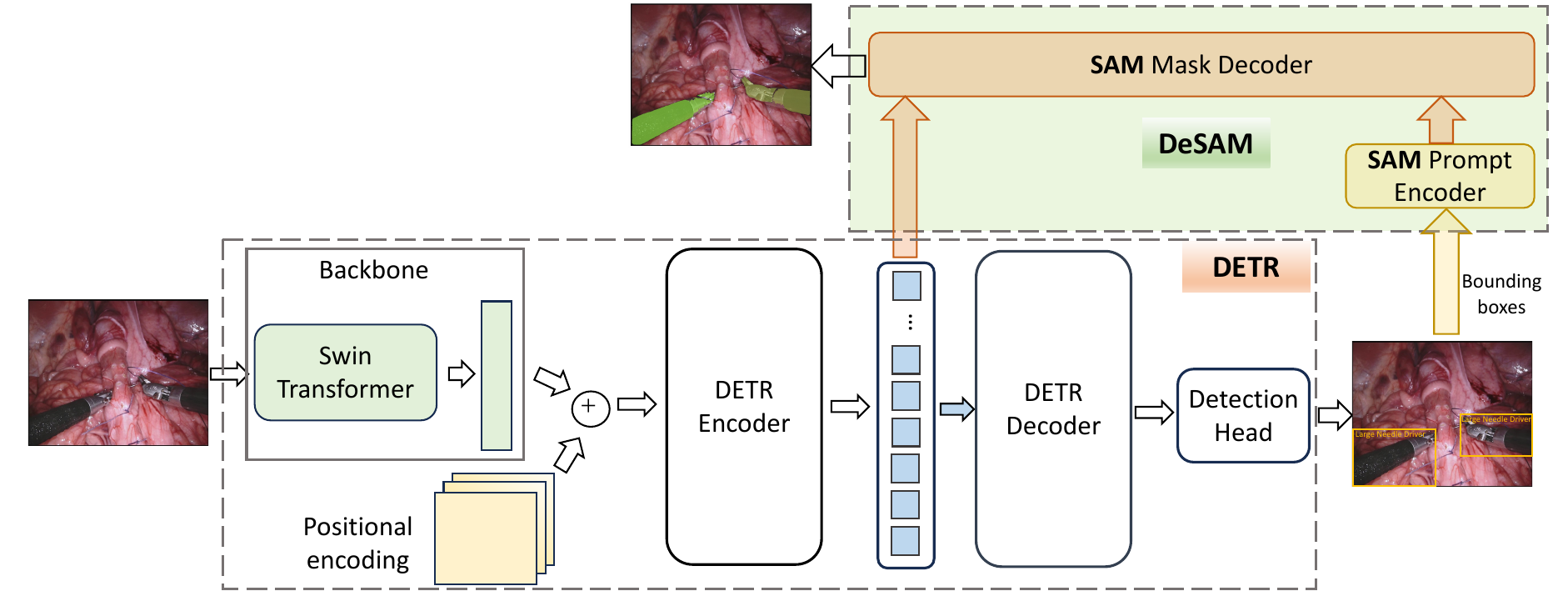}\label{fig:model_arch}
\caption{Surgical-DeSAM: Swin-DETR detector and Decoupling SAM for instrument segmentation.}
\end{figure}

\section{Methodology}

\subsection{Preliminaries}
\subsubsection{SAM}
SAM~\cite{kirillov2023segment} is the foundation model for prompt-based image segmentation and is trained on the largest segmentation dataset with over 1 billion high-quality masks. SAM forms a simple-designed transformer and composed of a heavyweight image encoder, a prompt encoder, and a lightweight mask decoder. The image encoder can directly extract image features from input images without the need for a backbone model, while its lightweight prompt encoder can dynamically transform any given prompt into an embedding vector in real-time. These embeddings are then processed by a decoder, generating precise segmentation masks. Prompts have various types, including points, boxes, text, or masks, which limit the SAM's ability to be directly utilised for real-world applications like surgical instrument segmentation during surgery. It is unrealistic to provide a prompt for each frame of the surgical video.

\subsubsection{DETR}
DETR~\cite{carion2020end} is the transformer-based detector called DETR (DEtection TRansformer) for object detection. It consists of a CNN backbone, an encoder-decoder transformer and feed-forward networks (FFNs). The CNN backbone is the commonly used ResNet50~\cite{he2016deep} which extract the feature ($\in\Re^{d\times H\times W}$) representation from the input image ($\in\Re^{3\times H_0\times W_0}$). The output of the backbone then passes to the transformer encoder with spatial positional encoding and produces object queries and encoder memory. The decoder receives the encoder outputs and predicts the class labels and bounding boxes with centre coordinates, height and width using FFNs.

\subsection{Surgical-DeSAM}
As shown in Fig.~\ref{fig:model_arch}, we proposed \textit{Surgical-DeSAM} to automate the bounding box prompting by designing (i) Swin-DETR: replacing ResNet50 of the DETR with Swin-transformer to design an efficient model for surgical instrument detection; (ii) Decoupling SAM: Replacing SAM image encoder with DETR Encoder and training end-to-end detection to prompt mask decoder of the SAM to segment surgical instrument.

\subsubsection{SWIN-DETR}
DETR utilises ResNet50 as the backbone CNN to extract the feature representation. However, as vision-transformer-based networks are showing much better performance than CNN,  we replace the backbone network with a recent transformer-based architecture of Swin-transformer~\cite{liu2021swin} and from our Swin-DETR as presented in Fig.~\ref{fig:model_arch}. The Swin-transformer introduces a shifted window-based hierarchical Transformer to add greater efficiency in the self-attention computation. It is important to note that the output of the Swin-transformer can be directly fed to the DETR encoder, where there is an additional step to collapse the spatial dimension of the ResNet50 feature into one dimension to convert it into a sequence of input for the transformer. Overall, SWIN-DETR consists of a Swin-transformer to extract the image feature, which is then passed to the transformer encoder-decoder and FFNs to obtain the final object class predictions and corresponding bounding boxes. More specifically, ResNet5o requires to convert feature map of $f_{resnet} \in\Re^{d\times H\times W}$ into $f \in\Re^{d\times HW}$ by collapsing the spatial dimensions where Swin-transformer directly produces output feature map of $f_{swin} \in\Re^{d\times HW}$.

\subsubsection{Decoupling SAM}
As the image encoder of the SAM and the DETR are performing similar feature extraction, we decouple SAM by removing the image encoder and feeding the DETR encoder output directly to the mask decoder. This facilitates the train end-to-end segmentation model using DETR predicted detection prompt and a decoupled SAM of prompt encoder and mask decoder only. During the training period, we utilise both ground-truths of the detection bounding boxes and segmentation masks to train both models end-to-end. To calculate losses, we adopted box loss $\mathcal{L}_{box}$ combining GIoU~\cite{rezatofighi2019generalized} and $l_1$ losses for the detection task following DETR and dice coefficient similarity (DSC) loss $\mathcal{L}_{dsc}$ for the segmentation task. Therefore, total loss $Loss_{total}$ can be formulated as:
\begin{equation}
\label{dscloss}
    Loss_{total} = \mathcal{L}_{box} + \mathcal{L}_{dsc}
\end{equation}

\section{Experiment and Results}
\subsection{Dataset}
We utilise two benchmark robotic instrument segmentation datasets of EndoVis17~\footnote{https://endovissub2017-roboticinstrumentsegmentation.grand-challenge.org/} and EndoVis18~\footnote{https://endovissub2018-roboticscenesegmentation.grand-challenge.org/}. The dataset consists of instrument segmentation for different video sequences. We split the EndoVis17 first video sequences of 1 to 8 for the training and the remaining sequences of 8 and 9 for the testing. For EndoVis18, we split the sequences of 2, 5, 9, and 15 for testing and the remaining sequences for training followed by ISINet \cite{gonzalez2020isinet}.
\subsection{Implementation Details}
 We choose AdamW optimizer with the learning rate of $10^{-4}$ and weight decay of 0.1 to update the model parameters. The baseline DETR and SAM codes are adopted from the official repositories which utilise Pytorch framework for deep learning network.

\subsection{Results}
\begin{figure}[!h]
    \centering
    \includegraphics[scale=0.4]{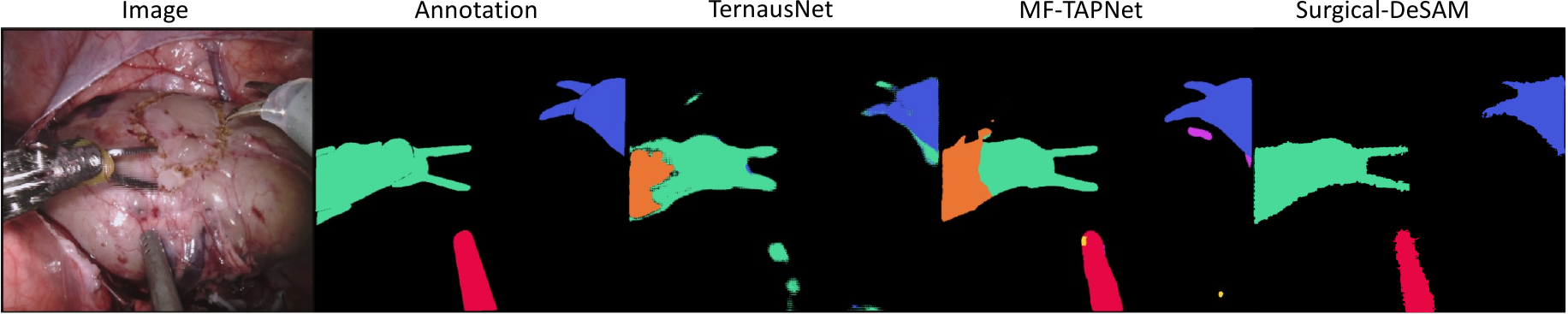}
    \caption{Comparison of the instance segmentation results with other models.}
    \label{fig:inst_seg_result}
\end{figure}

We conduct experiments on both object detection and semantic segmentation tasks on the robotic instrument dataset and obtain the instance segmentation performance of our model. Table \ref{table:overall_res} shows the comparison of performances of our model and other SOTA models for robotic instrument instance segmentation on Endovis 17 and Endovis 18 datasets. It is obvious that our Surgical-DeSAM outperforms the other SOTA segmentation models on both mIoU and DICE scores. The qualitative visualisation of the predictions is presented in Fig.~\ref{fig:inst_seg_result}. There are almost no false positives with our model as it segments the whole instrument based on the bounding box class predicted by the Swin-DETR. We observed the high detection performance with Swin-DETR at Table \ref{table:diff_bb} where predicted bounding boxes are mostly accurate with slight deviation of the box regions.

\begin{table*}[!t]
\centering
\renewcommand{\arraystretch}{1.2}
\setlength{\belowcaptionskip}{10pt}
\caption{Performance comparison of the proposed Surgical-DeSAM model and the SOTA models on EndoVis 2017 and 2018. Surgical-DeSAM outperforms significantly.}\label{table:overall_res}
    \centering
    \scalebox{.9}{
    \begin{tabular}{lccccc}
    \hline
        \multirow{2}{*}{Method} & \multicolumn{2}{c}{EndoVis 2017} & \multirow{2}{*}{Method} & \multicolumn{2}{c}{EndoVis 2018}\\
        ~ & mIoU & DICE & ~ & mIoU & DICE \\ \hline
        
        TernausNet \cite{iglovikov2018ternausnet} & 35.27 & - & TernausNet \cite{iglovikov2018ternausnet} & 46.22 & -\\
        MF-TAPNet \cite{jin2019incorporating} & 37.35 & - & MF-TAPNet \cite{jin2019incorporating} & 67.87 & -\\
        Dual-MF \cite{zhao2020learning} & 45.80 & 56.12 & Dual-MF \cite{zhao2020learning} & 70.41 & 76.93\\
        TrackFormer \cite{meinhardt2022trackformer}  & 54.91 & 59.72 & TrackFormer \cite{meinhardt2022trackformer} & 71.10 & 77.30\\ 
        ISINet \cite{gonzalez2020isinet} & 55.61 & 62.8 & ISINet \cite{gonzalez2020isinet} & 73.10 & 78.30 \\ 
        TraSeTR \cite{zhao2022trasetr} & 60.40 & 65.21 & TraSeTR \cite{zhao2022trasetr} & 76.20 & 81.10 \\ 
        S3Net (+MaskRCNN) \cite{baby2023forks} & 72.54 & - & S3Net (+MaskRCNN) \cite{baby2023forks} & 75.81 & -\\
        - & - & - & SurgicalSAM \cite{yue2023surgicalsam} & 80.33 & -\\
        - & - & - & Wang et al. \cite{wang2023sam} & 71.38 & -\\
        \textbf{Surgical-DeSAM} & \textbf{82.41} & \textbf{89.62} & \textbf{Surgical-DeSAM} & \textbf{84.91} & \textbf{90.70}\\ \hline
    \end{tabular}}
\end{table*}

\subsection{Ablation Study}
To investigate the superiority of the Swin-transformer \cite{liu2021swin} backbone over ResNet50 \cite{he2016deep}, we conducted an ablation study focusing on detection tasks alone and on both detection prompt and segmentation tasks. In Table \ref{table:diff_bb}, the first two rows demonstrate superior detection performance of DETR-SwinB (DETR with Swin-transformer) compared to DETR-R50 (DETR with ResNet50). Conversely, the subsequent rows compare the results of Surgical-DeSAM with ResNet50 and Swin-transformer backbones. It is evident that Surgical-DeSAM with a Swin-transformer backbone significantly outperforms Surgical-DeSAM with a ResNet50 backbone, achieving a 2.7\% higher mAP in the detection task and a 7.1\% higher DICE score in the segmentation task.

\begin{table}[!h]
\renewcommand{\arraystretch}{1.2}
\setlength{\belowcaptionskip}{10pt}
\caption{Comparison of DETR and our model with different backbone networks for the detection and segmentation tasks.}\label{table:diff_bb}
    \centering
    \begin{tabular}{cccccc}
    \hline
        \multirow{2}{*}{Method} & \multicolumn{3}{c}{Detection} & \multicolumn{2}{c}{Segmentation} \\ 
        ~ & mAP@0.50:0.95 & mAP@0.50 & mAP@0.75 & mIoU & DICE \\ \hline
        DETR-R50 & 61.4 & 82.6 & 71.3 & - & - \\ 
        DETR-SwinB & 64.6 & 83.4 & 73.1 & - & - \\ 
        Surgical-DeSAM (ResNet50) & 58.9 & 80.6 & 66.9 & 75.2 & 82.5 \\ 
        Surgical-DeSAM (Swin) & \textbf{61.6} & \textbf{83.2} & \textbf{71.2} & \textbf{82.4} & \textbf{89.6} \\ \hline
    \end{tabular}
\end{table}

\section{Discussion and Conclusion}
In this paper, we have presented a novel model architecture, Surgical-DeSAM, by decoupling SAM to automate the bounding box prompting for surgical instrument segmentation. To get better feature extraction, we replaced ResNet50 with the Swin-transformer for instrument detection. To automate the bounding box prompting, we decouple the SAM by removing the image encoder and feeding the DETR encoder features and predicted bounding boxes to the SAM mask decoder and prompt encoder to obtain the final segmentation. The experimental results demonstrate the efficiency of our model by comparing it with other state-of-the-art segmentation techniques for surgical instrument segmentation. Future work could focus on the robustness and reliability of the Surgical-DeSAM-based detection and segmentation tasks.

\section{Acknowledgements}
This work was carried during the dissertation project of Yuyang Sheng MSc in Robotics and Computation, Department of Computer Science, University College London. This work was supported in whole, or in part, by the Wellcome/EPSRC Centre for Interventional and Surgical Sciences (WEISS) [203145/Z/16/Z] and the Engineering and Physical Sciences Research Council (EPSRC) [EP/W00805X/1, EP/Y01958X/1].

\section{Declarations}
\textbf{Conflict of interest} The authors declare that they have no conflict of interest.

\noindent\textbf{Ethical approval} This article does not contain any studies with human participants or animals performed by any of the authors.

\noindent\textbf{Informed consent} This articles does not contain patient data.

\noindent\textbf{Code availability}
The source code of this work is available at \url{https://github.com/YuyangSheng/Surgical-DeSAM}.

\small \bibliography{references}

\end{document}